\documentclass{article}

\usepackage[nonatbib,preprint]{neurips2024}
\usepackage[backend=biber,style=numeric,natbib=true,maxcitenames=10]{biblatex}
\usepackage{graphicx}

\usepackage{listings}
\usepackage[utf8]{inputenc} 
\usepackage[T1]{fontenc}    
\usepackage{hyperref}       
\usepackage{url}            
\usepackage{booktabs}       
\usepackage{amsfonts}       
\usepackage{nicefrac}       
\usepackage{microtype}      
\usepackage[dvipsnames]{xcolor}         

\usepackage{amsthm}
\usepackage{amsmath}

\theoremstyle{definition}
\newtheorem{definition}{Definition}[section]

\title{Uncovering Latent Memories: Assessing Data Leakage and Memorization Patterns in Frontier AI Models}

\author{%
   Sunny Duan\\
   Brain and Cognitive Sciences\\
   MIT\\
   \texttt{sunnyd@mit.edu}
   \And 
   Mikail Khona\\
   Physics\\
   MIT\\
   \texttt{mikail@mit.edu}
   \And
   Abhiram Iyer\\
   EECS\\
   MIT\\
   \texttt{abiyer@mit.edu}
   \And
   Rylan Schaeffer\\
   Computer Science\\
   Stanford University\\
   \texttt{rschaef@cs.stanford.edu}
   \And
   Ila Rani Fiete\\
   Brain and Cognitive Sciences\\
   MIT\\
   \texttt{fiete@mit.edu}\\
}

\begin{document}

\maketitle

\begin{abstract}
 Frontier AI systems are making transformative impacts across society, but such benefits are not without costs: models trained on web-scale datasets containing personal and private data raise profound concerns about data privacy and security. Language models are trained on extensive corpora including potentially sensitive or proprietary information, and the risk of data leakage --- where the model response reveals pieces of such information --- remains inadequately understood. Prior work has investigated what factors drive memorization and have identified that sequence complexity and the number of repetitions drive memorization. Here, we focus on the evolution of memorization over training. We begin by reproducing findings that the probability of memorizing a sequence scales logarithmically with the number of times it is present in the data. We next show that sequences which are apparently not memorized after the first encounter can be ``uncovered'' throughout the course of training {\em even without subsequent encounters}, a phenomenon we term ``latent memorization". The presence of latent memorization presents a challenge for data privacy as memorized sequences may be hidden at the final checkpoint of the model but remain easily recoverable. To this end, we develop a diagnostic test relying on the cross entropy loss to uncover latent memorized sequences with high accuracy.
\end{abstract}

\section{Introduction}
Frontier AI models are trained on vast web-scale datasets \citep{Touvron2023-ax,Gemini_Team2023-yo,OpenAI2023-wv,Brown2020-gr}. The sizes of these pretraining corpora enable fluency, knowledge about various domains\citep{AlKhamissi2022-qe,Guu2020-cn}, and the ability to perform in-context learning \cite{Brown2020-gr}. However, these datasets often include proprietary, copyrighted, or otherwise private information \citep{Smith2023-tk,Karamolegkou2023-bv,Bordt2024-rm,Duan2024-vu,Staab2023-vu,Shi2023-do,Tang2023-ke,Zanella-Beguelin2019-um}. LLMs are capable of not only using training data for general knowledge and performance, but have been shown to possess a vast capacity for detailed memorization. Specifically, with appropriate prompting, LLMs can regurgitate verbatim text from their training corpii. With appropriate prompts, LLMs can reproduce verbatim text from their training data, a phenomenon that contrasts with ``catastrophic forgetting'' \citep{Kirkpatrick2017-xv,Zenke2017-qk,De_Lange2019-wr,Serra2018-tb,Wang2023-yd,Schwarz2018-gt,Ritter2018-ek}, where models quickly forget previously learned examples if they are not re-encountered throughout training.

Ordinarily, model training aims to prevent forgetting, but there are scenarios where it is beneficial to ensure models do not reproduce exact training examples to safeguard the privacy and confidentiality of the data. In language modeling, the goal is to retain language skills and reasoning capabilities without memorizing individual examples. As frontier AI models are often trained on a single pass of the training corpus, they necessarily encounter distributional shifts during training. In spite of this, they have been observed to recall verbatim sequences encountered early in their training.  \citep{Biderman2023-zu}. 

One hypothesis is that memorized sequences appear multiple times within the corpus, allowing the network to reinforce and store this data in its weights. Our findings confirm that repeated sequences constitute the majority of the memorized content. However, we also demonstrate that even sequences encountered only once during training are memorized by the model and persist throughout the training process. Additionally, many memories that seem to be forgotten at one point in training are later recalled without additional exposure to those sequences. These latent memories pose significant privacy and copyright concerns, as they may not be easily detected using current methods.

\subsection{Contributions}

This study provides significant insights into the dynamics and mechanics of memorization in Frontier AI models, contributing to the broader understanding of data privacy and security within machine learning. Our primary contributions are as follows:

\begin{itemize}
    \item \textbf{Quantification of Memorization Susceptibility}: We systematically evaluate how the statistical characteristics of training data, specifically sequence complexity and repetition, influence the likelihood of memorization in language models. Our findings demonstrate that the probability of memorizing a sequence scales logarithmically with its repetition in the training data as well as the complexity of the sequence under consideration.

    \item \textbf{Stationarity of Memorized Sequences}: Through detailed analysis of training dynamics, we discover that the memorization status of sequences remains largely stationary after initial exposure, despite not being re-encountered. This indicates a fundamental property of the model's memory mechanism, where the state of memorized sequences is fixed and subsequent training only modifies the readout.
    \item \textbf{Latent Memorization and Recovery}: We identify the presence of "latent" memorized sequences, which are not evident at certain checkpoints but can be uncovered later in training or through controlled perturbations. Our experimental results show that adding random Gaussian noise to model weights can recover these latent memorized sequences, supporting the hypothesis that further training acts as random additive noise rather than fundamentally altering the memorization state.

    \item \textbf{Development of a Diagnostic Test}: We propose a novel diagnostic test for uncovering latent memorized sequences by analyzing their cross-entropy loss. This test provides a practical tool for detecting and mitigating potential data leakage in deployed language models.
\end{itemize}

 Our study underscores the risks associated with data leakage in language models, emphasizing the need for robust mechanisms to ensure data privacy. The persistence of memorized sequences poses a challenge for the prevention of data leakage. By characterizing the nature of memorization as well as the nature of these latent memorized sequences, we elucidate possible mechanisms of how sequences become memorized and offer practical solutions for mitigating data privacy risks, and developing safer and more secure models.

\section{Methodology}
\setlength{\fboxsep}{0pt}
\setlength{\fboxrule}{0pt}

\subsection{Properties of Pretraining Data Relevant to Memorization: Repeats and Complexity}

Previous studies have identified that the number of repeats of a string affects whether it will be memorized, with more frequently occurring strings being more likely to be memorized \citep{Carlini2020-mz,razeghi2022impactpretrainingtermfrequencies, Biderman2023-ly}. Consequently, a starting property to measure is the \textbf{number of repeats} of specific strings in the pretraining corpus.

In our work, we also consider a second and newer property: the \textbf{complexity} of specific strings. Our decision to do so is motivated by previous studies which identified a prevalent class of easily memorized data: simple sequences composed of repeated patterns, numbers, or other straightforward patterns \cite{Carlini2020-mz}. While models readily learn these samples, they often lack substantive content and are unlikely to represent sensitive information. Thus, it is important to distinguish between memorization of these trivial sequences from more complex and informative sequences.

In order to quantitatively measure the complexity of specific strings, we turn to a widely used notion of informativeness called Kolmogorov complexity \cite{kolmogorov1963tables,kolmogorov1998tables}. Kolmogorov complexity is defined as the minimum description needed to specify a sequence. While this formalism is helpful in defining complexity, it is a theoretical measure that cannot be readily computed. As a proxy, we use modern compression algorithms to determine the extent to which sequences have a smaller description than the original sequence. To calculate the complexity of a sequence we define a metric, z-complexity, which is the ratio between the compressed sequence length to the original sequence length. 
This metric contains values from \(0\) to \(1\) and is efficiently computable using the zlib package in python. This metric is an upper bound on the Kologomorov complexity, since the Kologomorov complexity is defined as the smallest of such descriptions. 

\subsection{Quantifying memorization}

Several different definitions have been put forward to quantify memorization in frontier AI models. Intuitively, a memorized sequence is a training sequence which can be reproduced given the right conditions. One popular definition of memorization is \textbf{\(kl\)-memorization} \cite{Carlini2022-dd}. \(kl\)-memorization is evaluated by considering a sequence of length \(k+l\). The first \(k\) tokens are presented to the model as context. The model is used to generate a continuation of length \(l\) via greedy (i.e., temperature $=0$ decoding). The model's continuation is compared to the "true" continuation, and a sequence is said to be \(kl\) memorized if the model's output exactly matches the true continuation.


While undoubtedly useful, we introduce and utilize a new metric for measuring memorization. \(kl\)-memorization is an overly strict such that even single-token deviations from the true continuation may cause us to misclassify a sequence as forgotten; in many cases, the model may make small errors such as inserting or modifying a single token. We identified and provide several examples in Table \ref{samples}.
In order to to be more robust to small changes in the learned sequence, we propose a modification of \(kl\)-memorization by introducing \textbf{kl-Levenshtein distance (kl-LD)}.
\\
\begin{definition}[kl-LD distance]
Let \( S = (s_1, s_2, \ldots, s_n) \) be a sequence of tokens. We denote the first \(k\) tokens as the context \( C = (s_{1}, \ldots, s_k) \) and the last \(l\) tokens as the target \( T = (s_{k+1}, \ldots, s_{k+l}) \). The model is provided these context tokens and produces a predicted continuation of \(\hat{T} = (m_1, \ldots, m_l)\). We define the kl-LD distance as the Levenshtein distance \cite{Levenshtein1965-ue}  between sequences \(T\) and \(\hat{T}\) where the Levenshtein distance is the minimum number of (token) insertions, deletions or substitutions that must be performed on \(T\) to obtain \(\hat{T}\).

\end{definition}

We find that this is a natural measure of memorization which also provides a range of values to provide more granular insight into the strength of the model's memory, akin to how continuous metrics have recently been used to improve evaluations of language models elsewhere \citep{schaeffer2023emergentabilitiesmirage,schaeffer2024predictingdownstreamcapabilitiesfrontier}. Throughout this study, we set \(k=32\) and compare the continuation of the model with the original sequence by computing the Levenshtein distance between the next \(64\) tokens.
\begin{table}[h]
\tiny
\caption{Model continuations at various stages in training for a few selected sequences which were complex and encountered only once during training. Minimum edits are highlighted such that character edits are highlighted in orange, deletions are highlighted in red and new characters are highlighted in green.}
\label{samples}
\centering
\begin{tabular}{p{0.15\linewidth}  p{0.18\linewidth} p{0.18\linewidth} p{0.18\linewidth} p{0.18\linewidth}}
\toprule
Context & True Continuation & Checkpoint 10000 & Checkpoint 15000 & Checkpoint 19000 \\
\hline
  .r001  Decision Letter 0  Silva  Daniel de Paiva  Academic Editor  © 2020 Daniel de Paiva & 
  2020  Daniel de Paiva  This is an open access article distributed under the terms of the  Creative Commons Attribution License  , which permits unrestricted use, distribution, and reproduction in any medium, provided the original author and source are credited.  20 Apr 2020  P& 
  \colorbox{red}{ Silva}  2020  Daniel de Paiva\colorbox{red}{ Silva}  This is an open access article distributed under the terms of the  Creative Commons Attribution License  , which permits unrestricted use, distribution, and reproduction in any medium, provided the original author and source are credited.  \colorbox{orange}{20} Apr 2020\colorbox{green}{ } \colorbox{green}{P}&
   Silva  2020  Daniel de Paiva Silva  This is an open access article distributed under the terms of the  Creative Commons Attribution License  , which permits unrestricted use, distribution, and reproduction in any medium, provided the original author and source are credited.  \colorbox{orange}{26 Feb} 2020 &
    Silva  2020  Daniel de Paiva Silva  This is an open access article distributed under the terms of the  Creative Commons Attribution License  , which permits unrestricted use, distribution, and reproduction in any medium, provided the original author and source are credited.  \colorbox{orange}{10 May} 2020 \\

\hline
992> por favor ayudenme para instalar DBDesigner <him> hay <BluesKaj>!es | Guest17992&
<ubottu> Guest17992: En la mayorÃxada de canales Ubuntu se comunica en inglÃ©s. Para ayuda en EspaÃ±ol, por favor entre en los canales \#ubuntu-es o \#kubuntu-es. <BluesKaj>!es |&
<ubottu> Guest17992: En la mayorÃ a de canales \colorbox{red}{de} Ubuntu se \colorbox{orange}{comunica} \colorbox{red}{sÃ³lo} en inglÃ©s. \colorbox{orange}{Para} \colorbox{red}{busca} ayuda en \colorbox{orange}{Espa}Ã±ol\colorbox{orange}{, por}
\colorbox{orange}{favor entre}\colorbox{red}{ra} en \colorbox{orange}{los can}\colorbox{green}{ales} \#ubuntu\colorbox{green}{-es o} \colorbox{green}{\#kubuntu}-es. \colorbox{orange}{<BluesKaj>}\colorbox{green}{!} es \colorbox{orange}{|}&
<ubottu> Guest17992: En la mayorÃ a de canales de Ubuntu se habla sÃ³lo en inglÃ©s. Si busca ayuda en espaÃ±ol o charlar entra en el canal \#ubuntu-es. Escribe "/join \#ubuntu-es"&
<ubottu> Guest17992: En la mayorÃ a de\colorbox{green}{ los} canales de Ubuntu\colorbox{green}{,} se habla sÃ³lo en inglÃ©s. Si busca ayuda en espaÃ±ol\colorbox{orange}{ entre al} \colorbox{red}{lar entra en el} canal \#ubuntu-es\colorbox{orange}{; escriba} \colorbox{green}{ " /}join \#ubuntu-es\colorbox{orange}{ "}\colorbox{green}{ (} \\

\hline
 , findings, beliefs, or experiences on those topics or products. The views and opinions expressed on CateTheOkay.com are purely my own. Any product&
   claim, statistic, quote or other representation about a product or service should be verified with the manufacturer, provider or party in question.  CateTheOkay.com doesn’t contain any content which might present a conflict of interest.&
   claim, statistic, quote or other representation about a product or service should be verified with the manufacturer\colorbox{orange}{ or} provider\colorbox{orange}{.  Comments}\colorbox{red}{.}  \colorbox{orange}{I have a question}. \colorbox{orange}{I have a friend who is a} \colorbox{orange}{teacher and she} \colorbox{green}{is} a \colorbox{orange}{teacher. She is a teacher } \colorbox{orange}{and she is a student.} \colorbox{orange}{She is a student and she} \colorbox{orange}{is a}&
    claim, statistic, quote or other representation about a product or service should be verified with the manufacturer\colorbox{orange}{ or} provider\colorbox{red}{ or party in} \colorbox{red}{question}. CateTheOkay.com \colorbox{orange}{is} \colorbox{orange}{not affiliated with,} \colorbox{orange}{endorsed} \colorbox{orange}{by, or} \colorbox{orange}{sponsored} \colorbox{orange}{by the Coca}\colorbox{green}{-Cola Company}. \colorbox{orange}{CateTheOkay.com is not} \colorbox{orange}{affiliated with, endorsed by,}&
     claim, statistic, quote or other representation about a product or service should be verified with the manufacturer\colorbox{orange}{ or} provider\colorbox{red}{ or party in} \colorbox{red}{question}. \colorbox{orange}{I am not a doctor,} \colorbox{orange}{pharmacist, or registered} \colorbox{orange}{dietitian. I} \colorbox{green}{am not} a \colorbox{orange}{registered dietitian}. \colorbox{orange}{I am not a registered dietitian.} \colorbox{orange}{I am not a registered dietitian.} \colorbox{green}{I am}
\end{tabular}

\end{table}
\subsubsection{Analyzing Repeated Samples}
In this study, we seek to understand both how repeated encounters of a sequence during training drives memorization and also how sequences which are encountered only once are retained by the model. To this end, we analyze where training sequences are repeated throughout the course of training. In our study, we focus on the \(l\) portion of the sequence. For this study, we fixed \(l\) to be 64 tokens. Given this target sequence, we compare the target sequence with all of the training sequences which were presented to the model during the period of training under consideration. We compute the largest subsequence match between the target and every individual training example and call a training example a "repeat" if there was a sub-sequence match of length \(30\) or longer. We employed a parallelized data pipeline to search for repeats of 512,000 such target sequences.

\subsection{Language Models}
In this study, we largely focused on the Pythia 1B language model \cite{Biderman2023-ly}, which was trained on 300B tokens from the Pile \cite{Gao2020-pr}. For selected experiments, to ensure our results hold on other language models, we reproduced our results using a larger and better performing model, Amber-7B \cite{Liu2023-yk}. We selected these two models as they were large, high performing models complete with fully reproducible data sequences and frequent checkpoints. As in previous works \cite{Biderman2023-ly}, all experiments were run with the models run with half precision and temperature $0$.

\subsubsection{Language Model Checkpoints}
In our analysis, we used checkpoints from every 1000 training steps between from step 10k-20k in Pythia-1B and every revision of Amber-7B, corresponding to roughly 1.7 million training examples between revision 100 to 110. These selections were 10 checkpoints from each model which represented a sizable portion of training. These were chosen to be offset from the beginning of training to avoid artifacts or initial transients from random initialization, learning rate warmup and other peculiarities from initial phases of training.

\subsubsection{Weight Perturbations of Language Models}
In a later experiment, we study how memorization is affected by small perturbations to the parameters of our language models.
Weight perturbations were performed by adding random Gaussian noise scaled to have standard deviation of \(2 \times 10^{-3}\). These perturbations are similar to the size of changes to the weights of the model during training (Figure S\ref{s5}). We perturbed each model with 200 random perturbations and selected the model which had the lowest Leevenshtein distance from the true continuation.

\section{Experimental results}

\subsection{Data Statistics Predict Memorization}

\begin{figure}[t]
    \centering
    \includegraphics[width=0.7\textwidth]{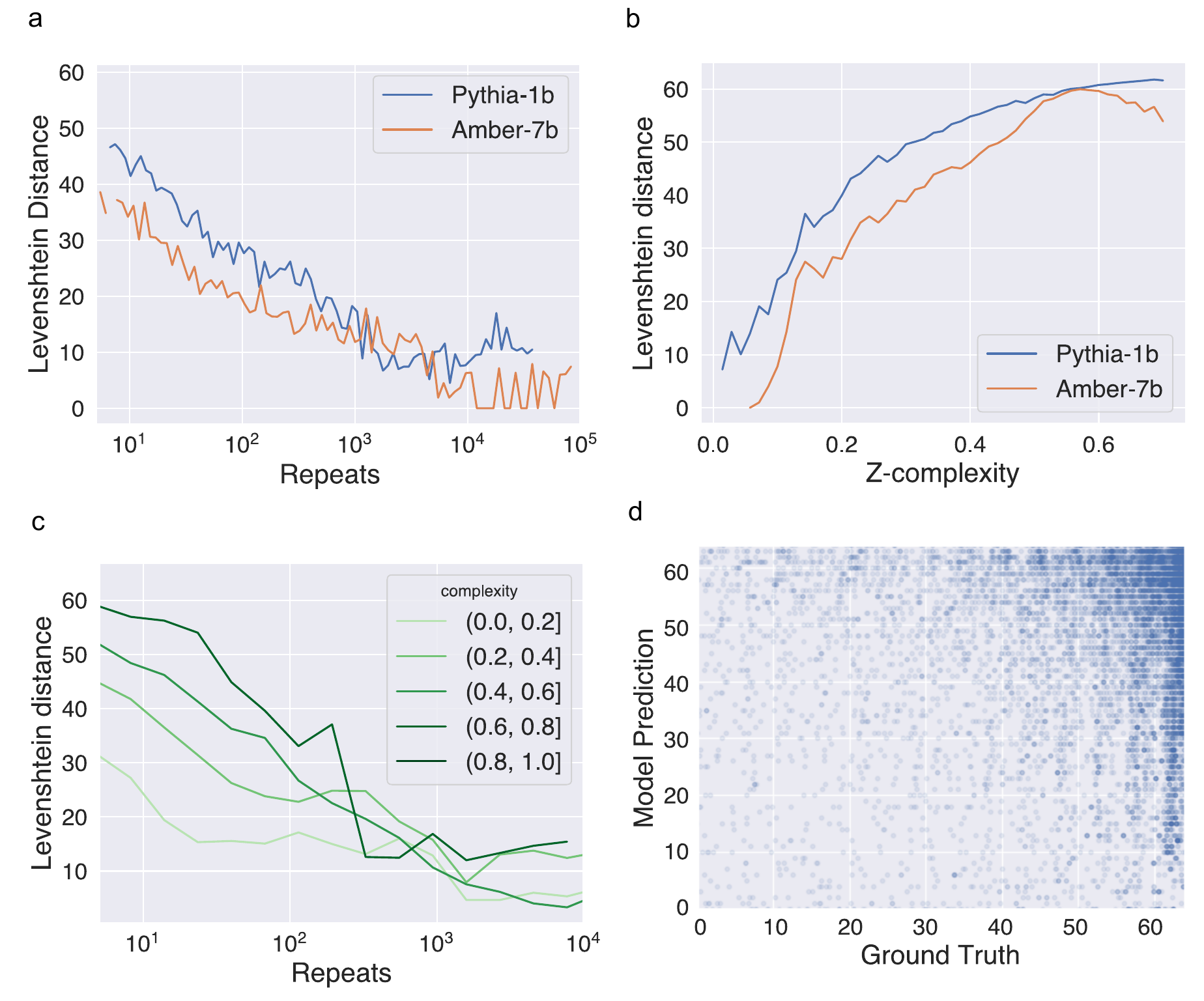}
    \caption{\textbf{Data statistics and the probability of memorization}
    \textbf{a.} Plot of average kl-LD as a function of the number of times the sequence is repeated in the dataset for Pythia-1b and Amber-7b \textbf{b.} Average kl-LD as a function of the Z-complexity of the sequence. \textbf{c.} Relationship between kl-LD and repeats for different complexity levels. \textbf{d.} Comparison of the predictions of the best linear model predicting the kl-LD from the logarithm of the sample complexity and number of repeats.
    }
    \label{fig1}
\end{figure}

We analyze two primary drivers of memorization during training: sequence complexity, and the number of repetitions. Previous work showed that the probability a training string can be extracted from a model is related to the model size and number of repetitions \cite{Carlini2020-mz}; we find that this relationship is true in the models we analyzed as well (Figure \ref{fig1}a). Additionally, we found that the complexity of the string itself was a strong predictor of whether it would be memorized (Figure \ref{fig1}b): strings with smaller z-complexity had smaller kl-Levenshtein distance (kl-LD), meaning simpler strings are more easily memorized. Interestingly, recent work showed that pretraining language models on data with lower z-complexity causes the training losses to decrease more rapidly \citep{pandey2024gzippredictsdatadependentscaling}; our results here suggest an explanatory mechanism: with more compressible data, the model can memorize the data more quickly. Furthermore, we found that for strings of different complexity exhibited different memorization curves (Figure \ref{fig1}c), whereby lower complexity strings were more easily memorized for smaller repeats. Both of these factors influenced the memorization probability with a log-linear relationship. 
While these factors were able to predict the probability of memorization, they did not fully determine whether a sequence will be memorized and significant uncertainty remains (Figure \ref{fig1}d). There are likely other factors which contribute to the memorization process such as the sequencing of training data and the state of the model when encountering the sequence.

\subsection{Dynamics of Memorization}

In order to produce a more complete picture of how successive training affects the state of memorized sequences within our model, we analyze how the kl-LD changes throughout the course of training for individual sequences. In order to eliminate the effects of repeated exposure to a string, we filter out sequences which are repeated throughout the course of training by eliminating sequences which are repeated according to our heuristic outlined above.
\begin{figure}[h]
    \centering
    \includegraphics[width=\textwidth]{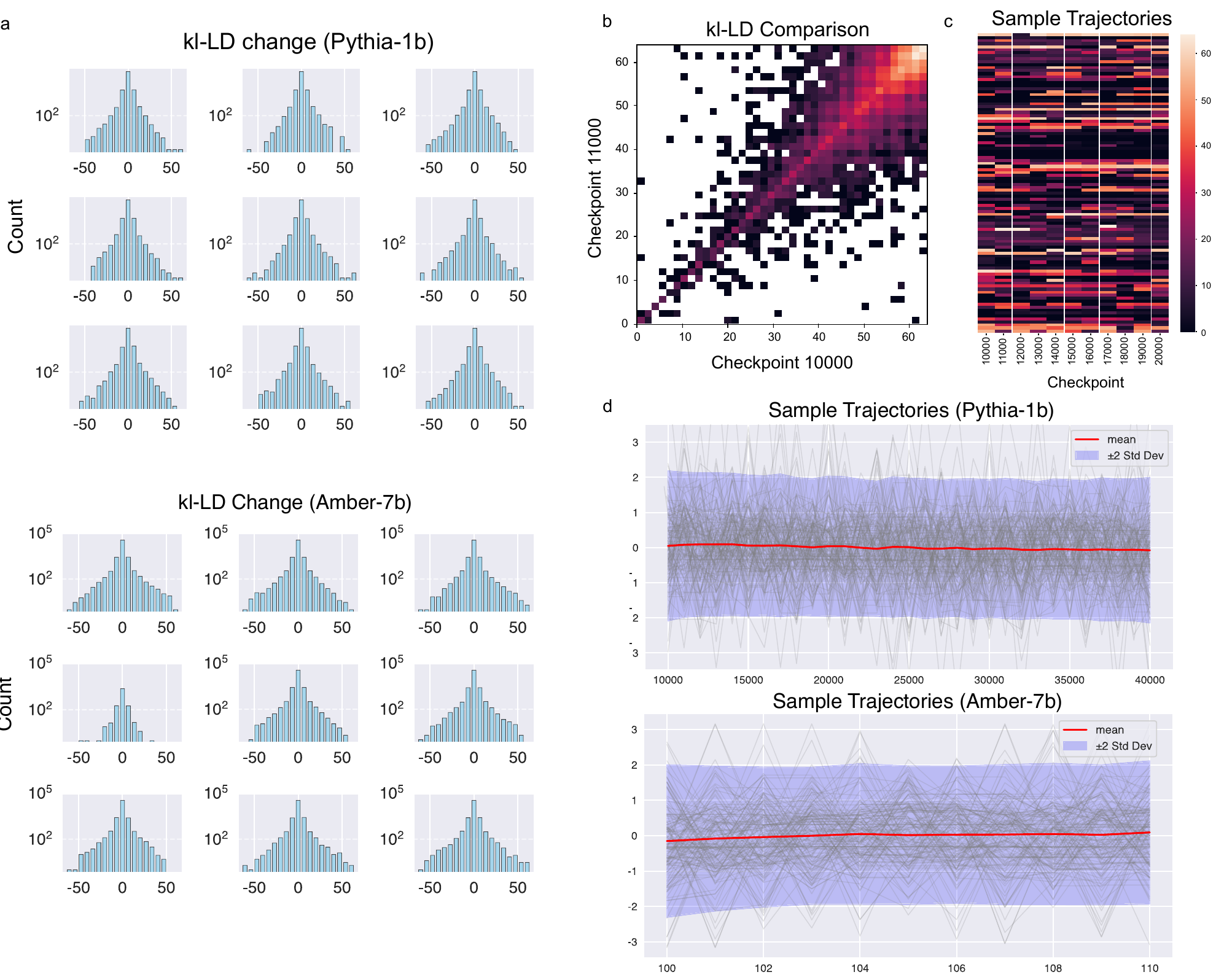}
    \caption{\textbf{Memorization status is stationary} a. Histograms of changes of edit distance between consecutive checkpoints for sequences which were encountered once during training. Notably, the change in kl-LD is symmetric between consecutive checkpoints. This is surprising since the model appears to "forget" the sequence during one timestep but recover it later on. 
    b. Distribution of kl-LD during checkpoint 10k and 11k. Color is the log of the number of sequences in each bin. The vast majority of sequences are not memorized in either checkpoint. 
    c. Visualization of individual samples and the change in the memorized length during training. 
    d. Grey lines are subsampled single sequence trajectories throughout training. Each sequence was normalized such that the distribution of memorization lengths was mean 0 and variance 1. Red line denotes the mean and shaded area denotes region of two standard deviations of the kl-LD of all sequences at a single point in time. Notably, the distribution at each timestep is the same for all checkpoints. This is in contrast to both the expected exponential decay behavior exhibited by models which experience catastrophic forgetting as well as the linear growth of variance which is expected of processes exhibiting random walk behavior.
    }
    \label{figure2}
\end{figure}
Surprisingly, we find that the memorization status of a sequence is largely stationary throughout training. After the initial checkpoint, the kl-LD of the sequences fluctuate but do so in a way which is stationary across training (Figure \ref{figure2}d). This is consistent across both Pythia-1b and Amber-7b models. This is reflected in the individual trajectories, and also in the overall mean of the population which shows no clear trend as training progresses. Furthermore, unlike a random walk, we see that the variance of the does not grow over time, but remains fixed. This is indicative of a mean reversion tendency of the dynamics and demonstrate the stability of the memories within the model weights. Additionally, we observe that the changes in the kl-LD between consecutive checkpoints (Figure \ref{figure2}ab) are symmetric and roughly follow a laplace distribution. This again confirms the counter-intuitive property of sequences to become memorized as often as they are forgotten. Notably, the model is able to recall memories which, at one point in time, appeared to be forgotten, despite never encountering that sequence again.

The stationarity of the memorization status of these sequences indicates that the memorized sequence is fixed throughout time, but this is in conflict with the fact that the model weights are constantly evolving. This stability in the presence of noise is indicative of a stabilizing mechanism by which the encoding of the sequence memory is preserved by some restorative process illustrated in Figure \ref{fig3}d where the memorized sequence becomes a fixed point in the weight space of the model under training dynamics. Subsequent training may alter the readout of the sequence, but the memory of the sequence is fixed throughout time. Since this is not true of all sequences, but only the few which exhibit this persistent memorization, it may point to a phase transition that occurs when the sequence is first encountered.

\subsection{Latent Memorization and Recovery of Latent Memories}
\begin{figure}[h]
    \centering
    \includegraphics[width=\textwidth]{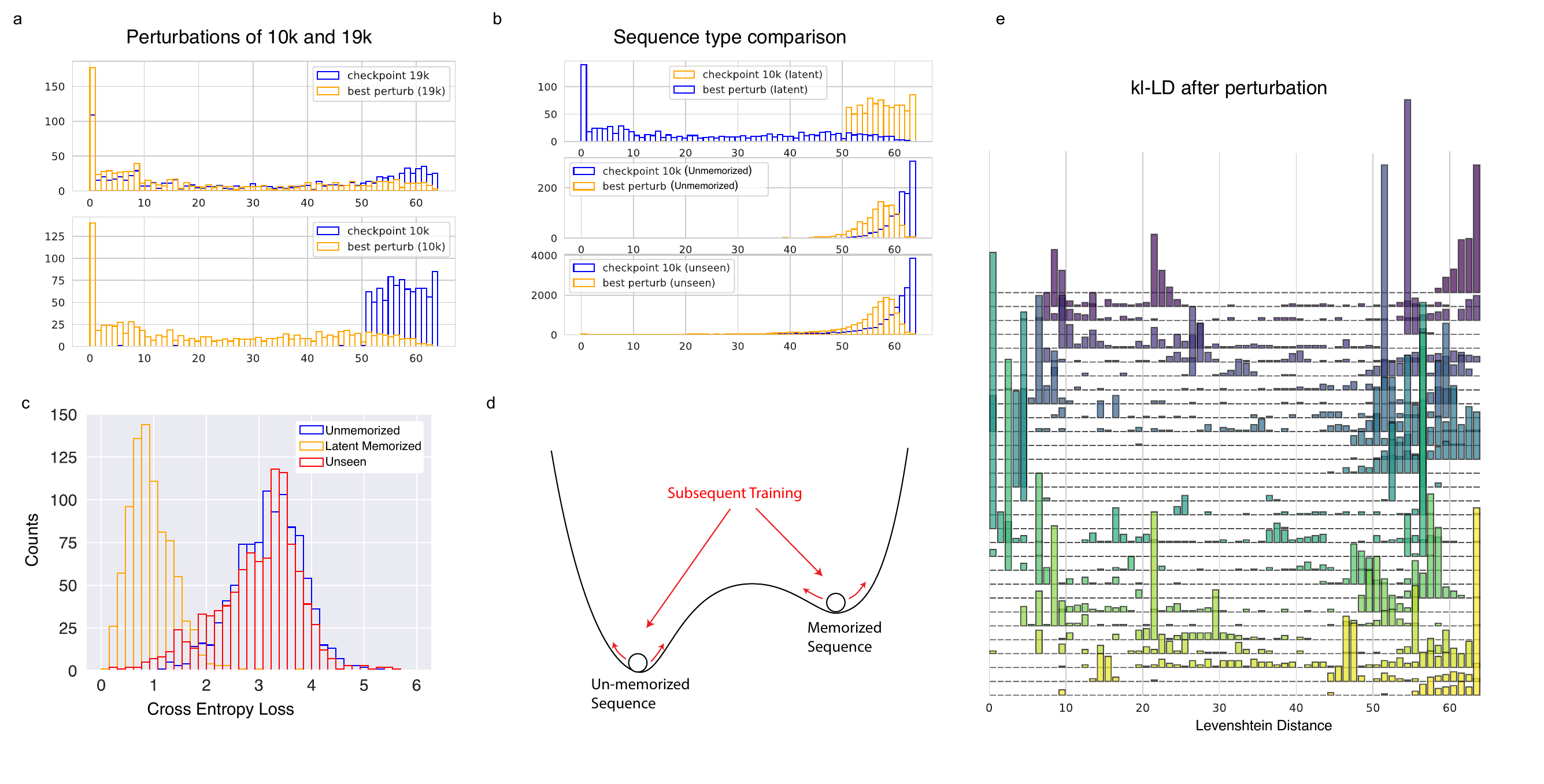}
    \caption{
    \textbf{a.} Comparison of the distribution of best achievable kl-LD by perturbing the model weights. Data points were selected such that they were un-memorized (kl-LD \(>\) 50) at 10k but we're memorized (kl-LD \(<10\)) at some point during the next 10k training steps. Top panel is the histogram of the perturbations of the checkpoint at 19k and bottom is 10k. Notably, the perturbations cause the 10k model distances to match the distribution of the 19k model, and perturbing the 19k model does not have a significant effect. This is indicative of how model training mimics random noise with respect to the memorization status of the sequences.
    \textbf{b.} Comparison of using perturbations to evoke a target sequence for three different classes of sequences. In the top panel, we examine the sequences which are "latent" memorized. In the middle panel, we find sequences which weren't memorized during training and in the bottom panel, we analyze sequences which were encountered later in training but were not encountered by the model. We not that perturbing the weights is only able to evoke sequences which are "latent" memorized.
    \textbf{c.} Comparison of the cross entropy losses of sequences separated into the three different classes of sequences analyzed in b. The cross entropy losses of "latent" memorized sequences are much lower.
    \textbf{d.} Drawing of a mechanistic proposal for how memorization is stabilized during training.
    \textbf{e.} Visualization of the Levenshtein distances from the target for various perturbations. Each row is a single sequence, and the heights of the bars correspond to the number of perturbations which resulted in a Levenshtein distance of the corresponding bin.
    }
    \label{fig3}
\end{figure}

Since some sequences exhibited seemingly random variations in their memorization state across different checkpoints, we hypothesize that these sequences remain memorized but are not be visible at a given checkpoint. Indeed, we found many sequences which were not memorized at the initial checkpoint (10k) but exhibited memorization by checkpoint 19k (Table \ref{samples}).

For these sequences, the nature of the random changes shown in Figure \ref{figure2} indicate the form of a random walk. We hypothesize that the process of training in frontier AI models acts as random noise on the weights with respect to the memory of the sequence. Thus, simply perturbing the weights with random noise should produce similar effects as training.

We find that this prediction is true. We randomly perturb the model weights by adding a small amount of random gaussian noise to each of the weight parameters. We repeat this process 200 times and find the perturbation which yields the lowest kl-LD. Notably, in the high dimensional weight space, it is difficult to reproduce arbitrary sequences using random weight perturbations, thus the recovery of memorized sequences must be due to intrinsic factors of how the memory is encoded in the weights.

We find that sequences which were not memorized at checkpoint 10k but were memorized later in training were able to be recovered using random perturbation (Figure \ref{fig3}a). In contrast, sequences which were not memorized during the period of consideration could not be recovered. As a control, we also selected sequences which were not presented to the model yet, and observed that their distributions closely matched those which were encountered by not memorized by the model (Figure \ref{fig3}b). Furthermore, we found that the perturbations yielded memorization patterns which closely matched that of the model at a later point in training. These observations support the view that with respect to a memorized sequence, subsequent training acts similar to random noise perturbations to the model weights. 

Finally, we find that these sequences which are not memorized at one point in training but appear later seem to be remembered by the model in spite of their incorrect continuation. These sequences can be considered to be "latent" memorized as they may not be visible at the current point in training, but they can be uncovered by small perturbations of the weights. These sequences pose a significant risk for leakage since they are not easily detectable from evaluating kl-memorization of those sequences. To this end, we discovered that these "latent" memorized sequences had significantly lower cross entropy loss when evaluated by the model (Figure \ref{fig3}c), thus simply evaluating the likelihood of those sequences using the trained model is a natural diagnostic for detecting these "latent" memorized sequences.

\subsection{Related Work}
Extracting memorized sequences from language models is an area of high interest. Early work established that it was possible to extract sensitive data including phone numbers, URLs and personal information from trained language models \citep{Carlini2020-mz}. Other studies injected canaries to determine what aspects of the training process contributed to whether a sequence is extractable \citep{Henderson2017-lv}\citep{Thakkar2020-fs}. More recent work have extended this to investigate how these properties scale with model size and data statistics \citep{Carlini2022-dd}. This has motivated the use of deduplication, which in addition to reducing the chance of data leakage \citep{Kandpal2022-ak}, also has been shown to improve sample efficiency and improve evaluation \citep{Lee2021-dd}. 

The definition of memorization is also still debated and various approaches to quantifying memorization have been made \citep{Zhang2021-yc,Feldman2020-bq}.
A variety of attacks have been designed to extract memorized sequences using designed prompts \citep{Thakkar2020-fs} and model activation perturbations \citep{Kassem2024-px}.

More generally, the notion of membership inference has been studied as a way to determine whether a given training example was part of the corpus \citep{Shokri2016-dh,Mireshghallah2022-df,Hisamoto2019-pg}, and these approaches have been applied to language models as well \citep{Duan2024-vu}.

Forgetting has also been studied extensively in neural networks, typically in the context of preventing forgetting. \citep{Kirkpatrick2017-xv,Zenke2017-qk,Chen2020-lq}. Studies have also shown that forgetting decreases with model size \citep{Tirumala2022-di,Mirzadeh2021-mr}. This work has also been examined in the context of understanding what aspects about a model and the data contribute to forgetting \citep{Toneva2018-tj}

Finally, there has also been work studying how the training process affects the status of memorization \citep{Tirumala2022-di}. This work focuses on how parameters of training and size of the model affect the dynamics of training. They find that scaling the model generally leads to less forgetting. In our work, we focus on sequences which counter-intuitively do not obey the forgetting laws presented in this work and expanding on the implications of these persistent "episodic" memories.

\section{Conclusion and Limitations}
We study how memorization changes throughout training and focused on sequences which occurred only once throughout training. Under these conditions, we find that rather than forgetting these sequences, the model retains them for the duration of training. This stationarity indicates a stability of the memorized sequence in weight space since the training process necessarily modifies the weights which encode the memorized sequences. We test this mechanistic view of how the training process interacts with the memorized sequence by using random weight perturbations to the model weights. These perturbations confirm that sequences which appeared to be forgotten at one point during training, may still be memorized by the model and are able to be uncovered with a small amount of random noise. We concluded by demonstrating a simple diagnostic to distinguish between "latent" memorized sequences and un-memorized sequences. 

This study highlights one surprising behavior of frontier AI models and begins to elucidate what mechanisms are present in the memorization behavior of these models. Our work suggest a possible mechanism of how memorized strings are sustained throughout training and further experiments are needed to confirm the underlying mechanism. Notably, further testing is required across other frontier AI models which were not considered here. We also propose a mechanistic explanation for this phenomenon which requires further study to explain the cause of these persistent memories. Finally, our analysis was restricted to a significant portion of training, but further analysis is needed to consider if these properties hold for even longer training durations.

\clearpage

\medskip
\printbibliography



\clearpage
\appendix

\section{Appendix / Supplemental Material}

\subsection{Compute details}
All experiments were run on a cluster with access to 16 concurrent a100 GPUs. All of the language models were run using a single GPU and multiple GPUs were used to parallelize the experiments in order to speed up progress. Searching for repeats within the dataset was performed using the library dask, using 64 CPUs distributed in a cluster, each with 32Gb of RAM.
\subsection{Licenses}
This project used code from the Pythia project \cite{Biderman2023-ly} released by EleutherAI under the Apache license version 2.0. We also used the Pile dataset \cite{Gao2020-pr} which is released under the MIT license. The Amber model was produced by LLM360, and the code and dataset are both released under apache 2.0.

\subsection{Additional figures}
We include figures which were ommitted from the main paper. These provide additional details that were not central to the claims made in the paper.

\begin{figure}[h]
    \centering
    \includegraphics[width=0.7\textwidth]{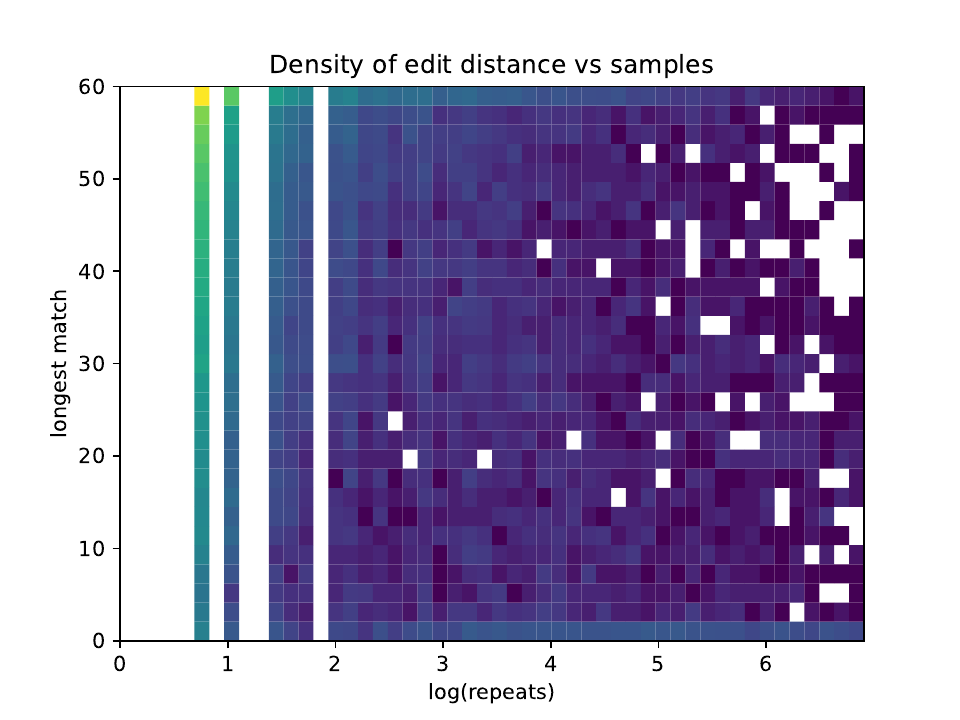}
    \caption{\textbf{Histogram of the repeats vs the edit distance} Hue is log density.
    }
    \label{s1}
\end{figure}

\begin{figure}[h!]
    \centering
    \includegraphics[width=\textwidth]{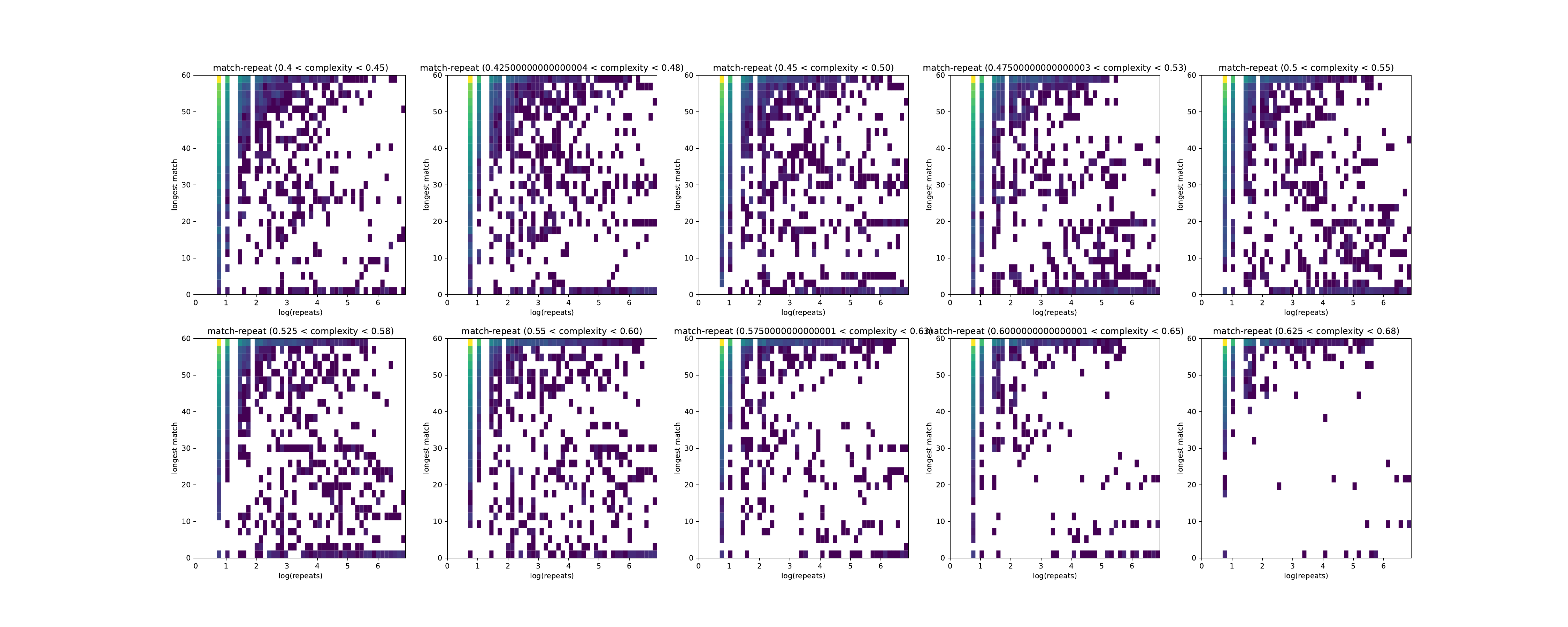}
    \caption{\textbf{Histogram of the repeats vs the edit distance split by complexity} Hue is log density.
    }
    \label{s2}
\end{figure}

\begin{figure}[h!]
    \centering
    \includegraphics[width=0.5\textwidth]{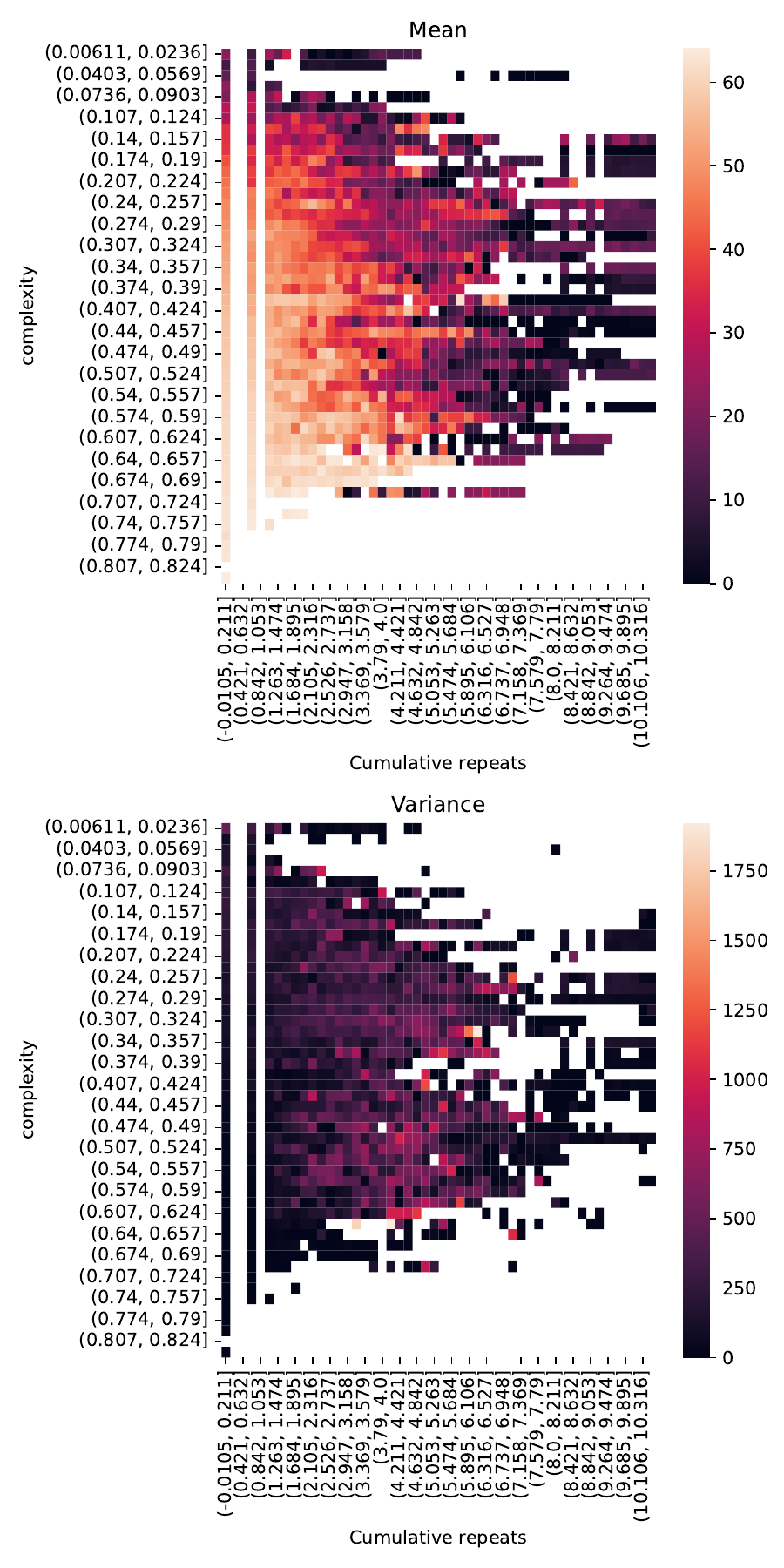}
    \caption{\textbf{Average of the kl-LD metric} kl-LD values are binned by number of repeats and complexity and the mean and variance of the samples in those bins are computed and colored.
    }
    \label{s3}
\end{figure}

\begin{figure}[h!]
    \centering
    \includegraphics[width=0.5\textwidth]{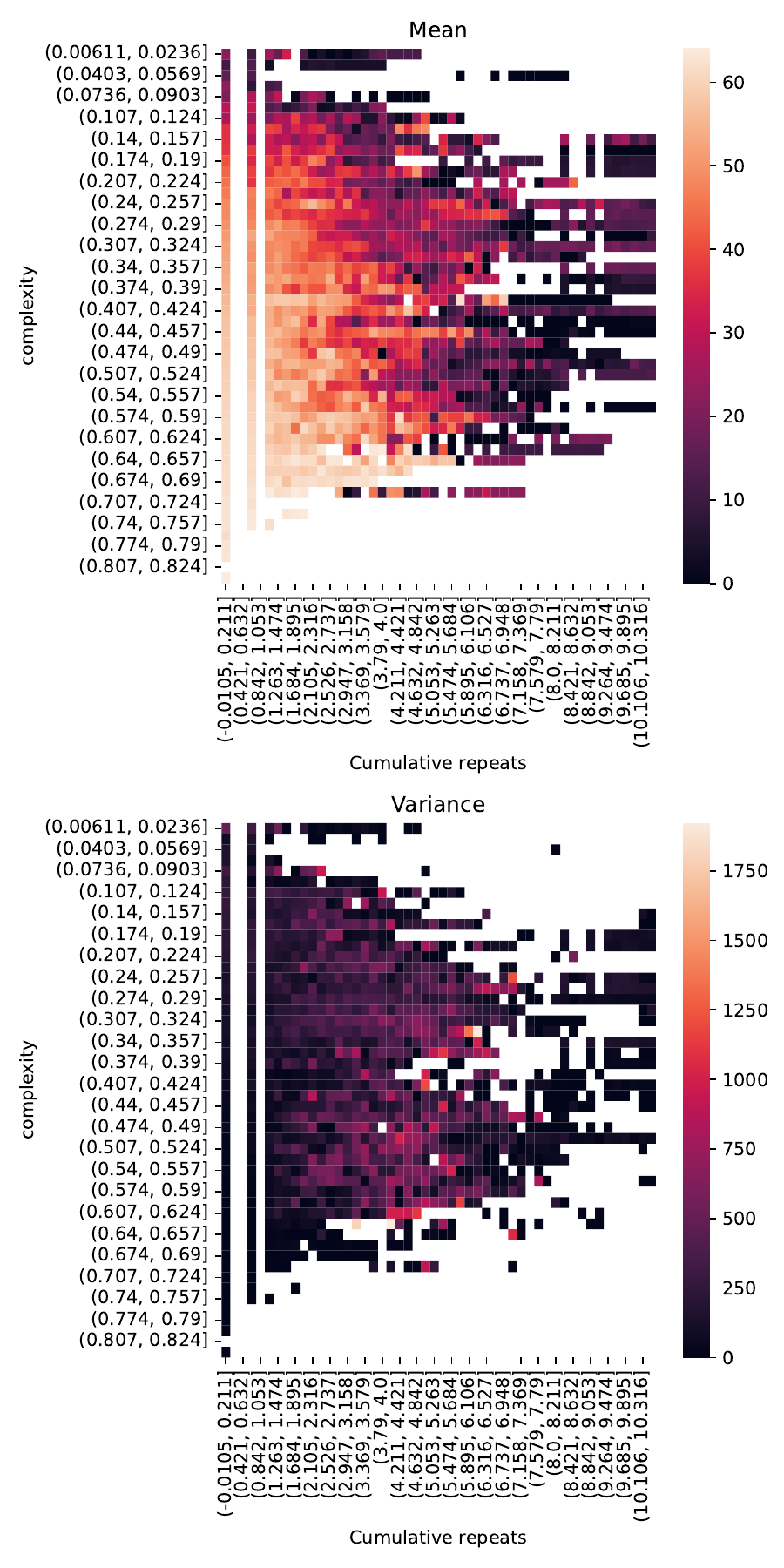}
    \caption{\textbf{Average of the kl-LD metric} kl-LD values are binned by number of repeats and complexity and the mean and variance of the samples in those bins are computed and colored.
    }
    \label{s4}
\end{figure}

\begin{figure}[h!]
    \centering
    \includegraphics[width=\textwidth]{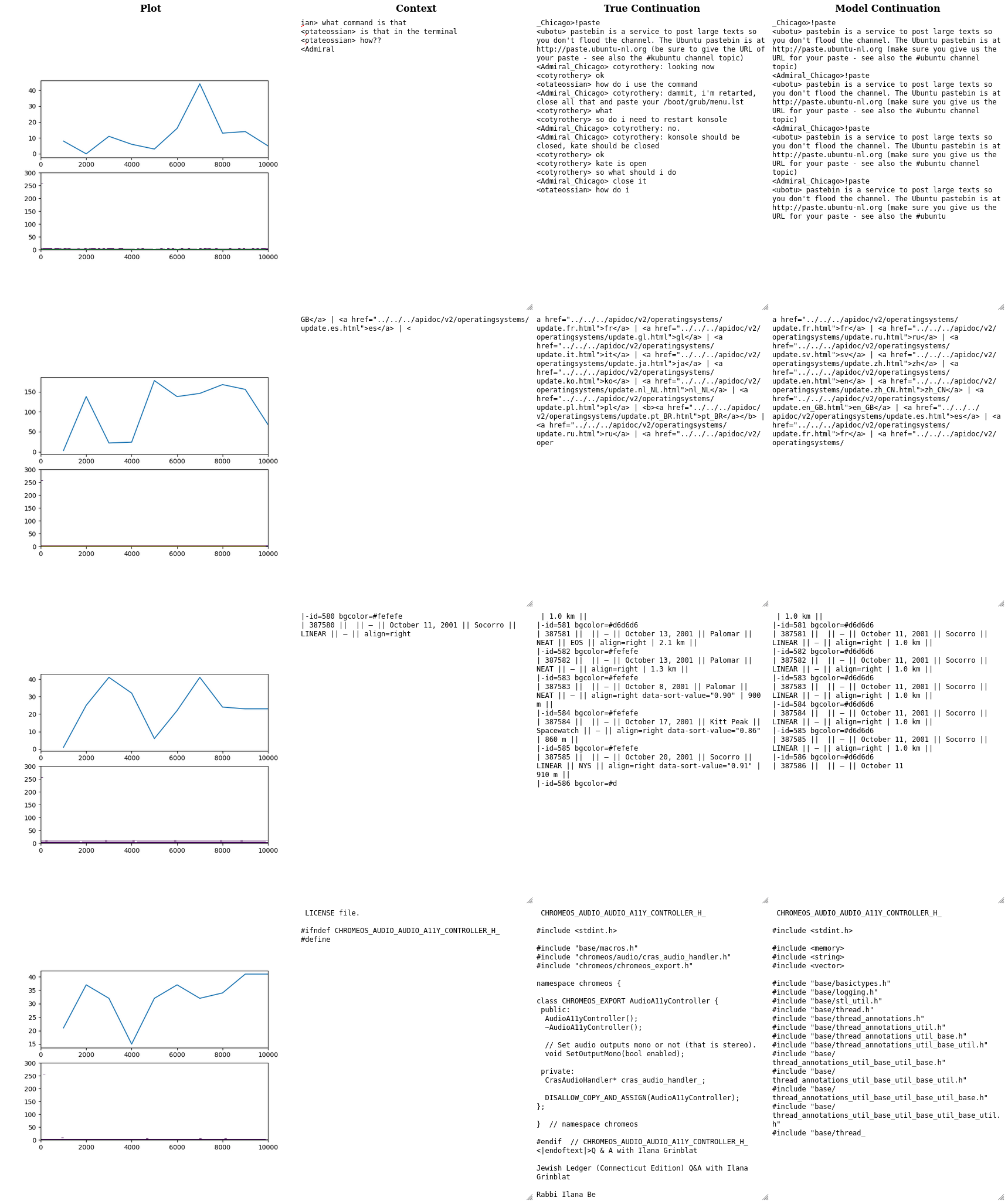}
    \caption{ Examples of strings which were seen once during training. Top left plot shows the kl-LD over for different trajectories and bottom left plot is a histogram of when the examples were repeated and at what length with the time on the x axis and the length of the repeat on the y axis. The text of the context, true continuation and model continuation are shown as well.
    }
    \label{s4}
\end{figure}

\begin{figure}[h!]
    \centering
    \includegraphics[width=0.5\textwidth]{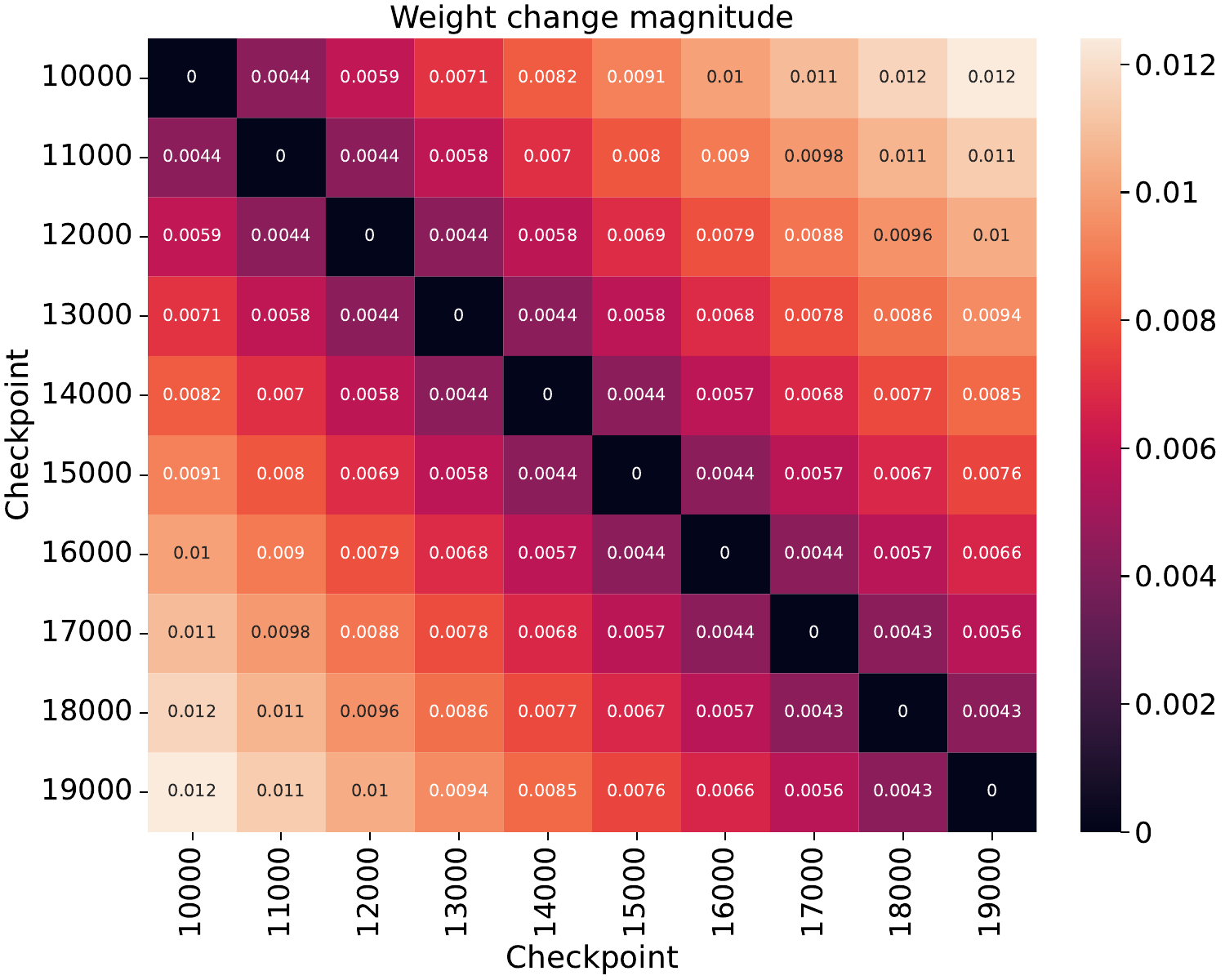}
    \caption{Distribution of weight changes of the model throughout training. Computed as the L2 distance between the flattened model weights at two different times during training
    }
    \label{s5}
\end{figure}
\begin{figure}[h!]
    \centering
    \includegraphics[width=0.5\textwidth]{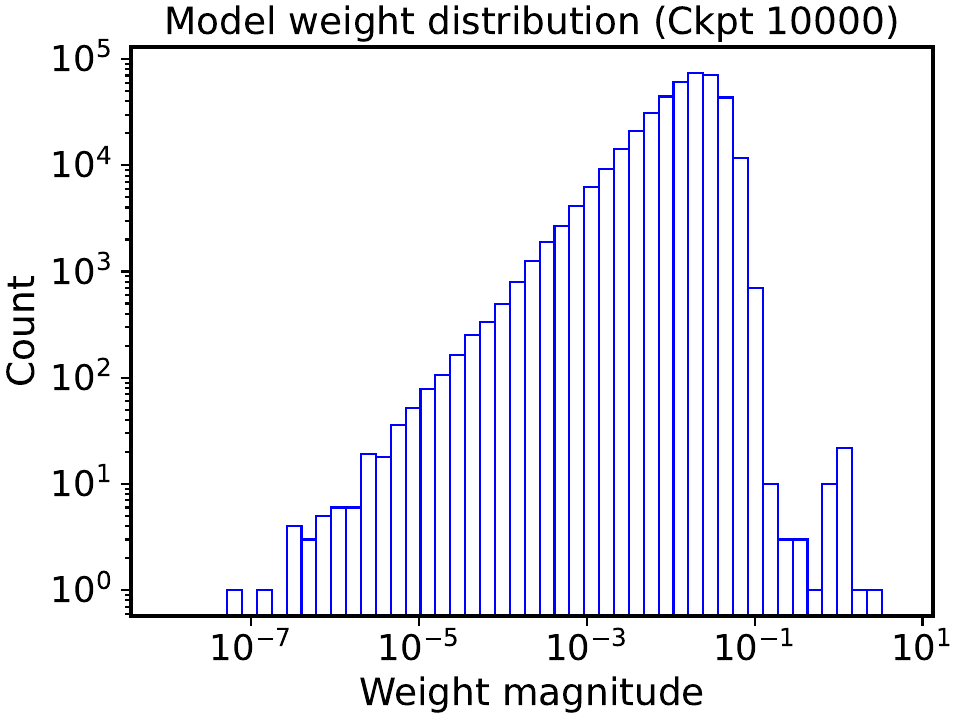}
    \caption{Distribution of weight magnitudes of the trained model at checkpoint 10k
    }
    \label{s6}
\end{figure}

\end{document}